\documentclass[sigconf,screen,nonacm]{acmart}

\usepackage{algorithm}
\usepackage{algorithmic}
\usepackage{mathtools}

\AtBeginDocument{%
  }

\begin{document}

\title{BayaHAR: Lightweight Bayesian Few-Shot User Adaptation for On-Device Personalized Human Activity Recognition}

\author{Maximilian Burzer}
\email{maximilian.burzer@kit.edu}
\orcid{0009-0000-9628-8667}
\affiliation{%
  \institution{Karlsruhe Institute of Technology}
  \city{Karlsruhe}
  \country{Germany}
}

\author{Till Riedel}
\email{till.riedel@kit.edu}
\orcid{0000-0003-4547-1984}
\affiliation{%
  \institution{Karlsruhe Institute of Technology}
  \city{Karlsruhe}
  \country{Germany}
}

\author{Michael Beigl}
\email{michael.beigl@kit.edu}
\orcid{0000-0001-5009-2327}
\affiliation{%
  \institution{Karlsruhe Institute of Technology}
  \city{Karlsruhe}
  \country{Germany}
}

\author{Tobias Röddiger}
\email{tobias.roeddiger@ipai-foundation.ai}
\orcid{0000-0002-4718-9280}
\affiliation{%
  \institution{IPAI Foundation gGmbH} %
  \city{Heilbronn}
  \country{Germany}
}
\renewcommand{\shortauthors}{Maximilian Burzer, Till Riedel, Michael Beigl, and Tobias Röddiger}

\begin{abstract}
Sensor-based Human Activity Recognition (HAR) models often degrade on unseen users due to domain shifts caused by individual movement patterns and sensor placement. Practical wearable HAR systems therefore require personalization methods that are lightweight, applicable across diverse calibration scenarios, and robust under limited calibration data. We present BayaHAR, a gradient-free framework that repurposes pretrained HAR classifiers as Prototypical Networks using prior prototypes that preserve zero-shot performance while regularizing adaptation. For labeled calibration data, we introduce closed-form Bayesian prototype estimation and extend the same principle to weakly labeled data, requiring only knowledge of which activities were performed. With only 3 seconds of calibration (one shot) per activity, supervised adaptation improves test macro-F1 on unseen users by +2.76 to +33.44 percentage points across four datasets, while weakly supervised adaptation improves by +0.56 to +32.13 points. Since adaptation requires only closed-form prototype updates, the framework enables efficient and robust on-device personalization of preexisting HAR classifiers.
\end{abstract}

\begin{CCSXML}
<ccs2012>
   <concept>
       <concept_id>10003120.10003138</concept_id>
       <concept_desc>Human-centered computing~Ubiquitous and mobile computing</concept_desc>
       <concept_significance>500</concept_significance>
       </concept>
    <concept>
       <concept_id>10010147.10010257.10010258</concept_id>
       <concept_desc>Computing methodologies~Learning paradigms</concept_desc>
       <concept_significance>500</concept_significance>
       </concept>
 </ccs2012>
\end{CCSXML}

\ccsdesc[500]{Human-centered computing~Ubiquitous and mobile computing}
\ccsdesc[500]{Computing methodologies~Learning paradigms}

\keywords{Human Activity Recognition; Domain Adaptation; Few-Shot Learning; Prototypical Networks; Bayesian Inference}

\maketitle

\section{Introduction}

Sensor-based Human Activity Recognition (HAR) models often degrade when deployed on unseen users due to domain shifts caused by inter-subject variability, including physiological differences in movement patterns and variations in sensor placement \cite{hossen2025machine, dhekane2025transfer}. Practical HAR systems therefore require adaptation mechanisms that remain effective across different calibration scenarios. At the same time, personalization should be lightweight enough for resource-constrained wearable devices, where latency, memory, and battery consumption limit the use of gradients \cite{lin2023tiny}.

To address these requirements, we build on Prototypical Networks \cite{snell2017prototypical}, an efficient framework for few-shot classification. However, conventional Prototypical Networks assume that a labeled support set is available during inference and therefore lack a natural zero-shot fallback when no calibration data is provided. Moreover, standard prototype estimation relies on empirical support means, which can become unstable in the ultra-low-shot regime due to sample noise and high variance \cite{yuan2021learning}. These issues are particularly problematic for HAR, where calibration should be as short as possible and weakly labeled recordings, with only knowledge on which activities were performed, are often more realistic \cite{sheng2025reducing}.

\begin{sloppypar}

We therefore introduce BayaHAR (Bayesian Adaptive HAR), a gradient-free framework that repurposes pretrained HAR classifiers as Prototypical Networks with source-domain prior prototypes. These priors preserve the original zero-shot behavior without calibration data and serve as regularizers when labeled or weakly labeled support samples are introduced. For labeled data, we update prototypes through closed-form Bayesian Maximum A Posteriori (MAP) estimation. For weakly labeled data, we combine this principle with an Expectation-Maximization (EM) update over latent activity class assignments. In both cases, the prior enables robust adaptation in low-support regimes while retaining a low computational overhead. We publicly release our code and experimental setup\footnote{\url{https://github.com/maxbrzr/baya-har}}. Our key contributions are summarized as follows:

\end{sloppypar}

\textit{(1) Repurposing via Prior Prototypes:} We introduce prior prototypes, which allow existing pretrained neural HAR classifiers to be repurposed as Prototypical Networks, while retaining their standard classifier behavior. This preserves zero-shot performance with negligible computational overhead and provides a unified starting point for few-shot adaptation.

\textit{(2) MAP Prototype Estimation:} We introduce closed-form MAP prototype estimation for labeled calibration data, regularizing noisy support estimates. With one shot (3 seconds) per activity, this improves macro-F1 on four datasets, with gains from \(+2.76\) to \(+33.44\) pp. With 16 shots (48 seconds), gains range from \(+6.8\) to \(+33.7\) pp.

\textit{(3) MAP-EM Prototype Estimation:} We extend prototype estimation to weakly labeled calibration data using single-step MAP-EM, requiring only knowledge about which activity classes are present. With one shot (3 seconds) per activity, MAP-EM improves macro-F1 over the zero-shot prior, with gains from \(+0.56\) to \(+32.13\) pp, while avoiding degradation observed in baselines. With 16 shots (48 seconds), gains range from \(+0.9\) to \(+32.5\) pp.

\vspace*{-.3em}

\section{Related Work}
\label{sec:domain_adapt}

In sensor-based HAR, deploying models on unseen users introduces significant domain shifts \cite{hossen2025machine, dhekane2025transfer}. These shifts primarily emerge from variations in sensor placement, which induce covariate shifts, and physiological differences in how subjects perform movements, which induce class-conditional shifts. While domain adaptation has been extensively researched to address these overlapping challenges, practical deployment on wearable devices introduces strict constraints that limit the applicability of many standard techniques.

Due to the strict memory, computational, and energy constraints of wearable devices, on-device user adaptation must be \textit{source-free}: relying solely on pretrained model parameters without requiring access to the original training data. This requirement disqualifies conventional domain adaptation methods that rely on source data, such as Domain-Adversarial Neural Networks (DANN) \cite{ganin2016domain}, and Deep CORAL \cite{sun2016deep}. Furthermore, on-device adaptation necessitates \textit{gradient-free} approaches. Backpropagation is computationally expensive and popular mobile machine learning frameworks like ExecuTorch \cite{pytorch_executorch} and TensorFlow Lite \cite{tensorflow_lite} are designed primarily for inference and lack native backpropagation support. This constraint eliminates parameter-efficient fine-tuning approaches like LoRA \cite{hu2022lora}, optimization-based meta-learning like MAML \cite{finn2017model} and prominent gradient-based test-time adaptation (TTA) methods such as SHOT \cite{liang2020we} and Tent \cite{wang2020tent}. While recent gradient-free TTA methods like OFFTA \cite{wangOptimizationFreeTestTimeAdaptation2024} and NEO \cite{murphyNEONoOptimizationTestTime2026} bypass backpropagation, they perform continuous online adaptation, which can be problematic on uncontrolled sensor streams, where updates may accumulate errors, reinforce noisy pseudo-labels, and cause catastrophic forgetting \cite{niu2022efficient, yang2024test}. These risks are particularly relevant in HAR, where incoming windows are often temporally correlated and low-diversity, so adaptation may overfit to recent activities or contexts rather than remain stable over longer sessions \cite{qiu2026coda}. Our work therefore targets a dedicated \textit{offline} calibration phase, using a limited set of user-specific calibration data, after which the adapted model is fixed to ensure stable personalization.

To enable \textit{source-free}, \textit{gradient-free}, and \textit{offline} adaptation, few-shot approaches rooted in Prototypical Networks \cite{snell2017prototypical} offer a promising direction. While originally supervised, unsupervised extensions such as PDA \cite{bohdalFeedForwardSourceFreeDomain2023} have also been developed. Within HAR, prototype-based methods have predominantly been explored in Federated Learning contexts \cite{rashidFederatedFewShotPrototypical2026, chengProtoHARPrototypeGuided2023, songPrototypeguidedPseudolabelingSemisupervised2026} or applied to continual learning scenarios for adapting to novel activity classes \cite{adaimiLifelongAdaptiveMachine2022a, huProtoCSNetPrototypeNetwork2024, jiangDualprototypeNetworkCombining2024}, which is beyond our scope. Recent works \cite{liBridgingDomainInstance2026, limPrototypeGuidedPhysicsModulatedPerceiver2026} have applied prototype-based adaptation to address domain shifts in HAR, however they are neither source- nor gradient-free. Moreover, all mentioned prototype-based methods fundamentally require support sets and cannot fallback to standard HAR classifiers when calibration data is unavailable.

Related few-shot paradigms such as self-supervised HAR foundation models also leverage reusable embedding spaces for few-shot transfer, but focus on broader cross-dataset, cross-modal, cross-placement, or language-grounded generalization \cite{bian2026foundation}. They are orthogonal to our work, as our adaptation strategy could be applied on top of such pretrained embeddings.

To the best of our knowledge, no existing approach fulfills all the mentioned requirements for practical offline user adaptation in HAR on wearable devices: being source-free, gradient-free, and computationally efficient, while flexibly and robustly handling zero-shot, supervised, and weakly supervised calibration scenarios. Building upon Prototypical Networks \cite{snell2017prototypical}, our work bridges this gap.

\section{Prototypical Networks}
\label{sec:proto_nets}

Prototypical Networks \cite{snell2017prototypical} (ProtoNets) constitute a few-shot and metric-based meta-learning framework that maps raw data into a learned $d$-dimensional continuous embedding space using an embedding network $f_\theta$. In this space, embeddings of samples belonging to the same class $k$, cluster around a single prototype representation $c_k$. Given a support set containing support examples $S_k$ for every class $k$, the prototypes are computed as the embedding mean:

\begin{equation}
    c_{k}=\frac{1}{|S_{k}|}\sum_{x\in S_{k}}f_{\theta }(x).
    \label{eq:proto_aggr}
\end{equation}

\begin{figure}[b]
    \centering
    \includegraphics[width=0.8\linewidth]{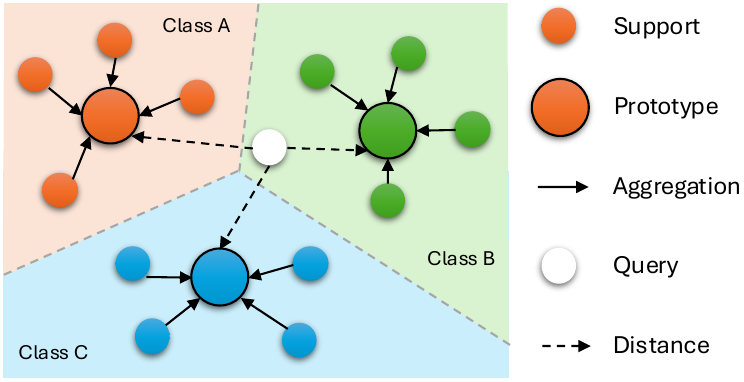} 
    \caption{During inference, the query sample is compared to three prototypes computed as the means of their respective support examples in embedding space. }
    \Description{TODO}
    \label{fig:proto}
\end{figure}

To classify an unseen query sample $x_q$, its embedding $f_\theta(x_q)$ is compared to each prototype $c_k$, as illustrated in \autoref{fig:proto}. For this purpose, the standard formulation uses the squared Euclidean distance:

\begin{equation}
    d(f_\theta(x_q),c_k)=\|f_{\theta }(x_q)-c_{k}\|^{2}.
    \label{eq:dist}
\end{equation}

The probability $p(y = k \mid x)$ of query sample $x$ belonging to class $k$ is then computed using a softmax over the negative distances to the prototypes. Consequently, the prototype with the smallest distance corresponds to the predicted class.

The embedding network $f_\theta$ is typically trained using episodic training \cite{snell2017prototypical}, a common strategy in meta-learning. During each episode, support and query sets are sampled in pairs and passed through the model, mimicking the few-shot setting and encouraging the model to learn how to generalize from limited labeled examples rather than memorizing large datasets.

\section{Methodology}
\label{sec:methodology}

In the following, we describe repurposing of ProtoNets with prior prototypes for when no calibration data is available (\autoref{sec:prototype_priors}). We then show how these priors can be updated with labeled (\autoref{sec:sup_proto_updating}) or weakly labeled (\autoref{sec:unsup_proto_updating}) support examples.

\subsection{Prior Prototypes}
\label{sec:prototype_priors}

While ProtoNets \cite{snell2017prototypical} were originally designed for few-shot learning to accommodate entirely novel classes, our work leverages their architecture for user adaptation instead. By restricting our scope to a fixed set of predefined activity classes, we eliminate the strict requirement for a target-domain support set during inference, unlocking significantly greater flexibility for real-world deployments. 

To enable classification when user-specific calibration data is unavailable, we propose defining a set of \textit{prior prototypes}. Instead of relying on a support set $S_k$, we compute a global prototype per class $k$ as the mean of all embedded training data points $D_k$, replacing the support examples in \autoref{eq:proto_aggr}. Therefore, this repurposing step itself is not \textit{source-free}, but can be performed offline before deployment. The resulting prior prototypes are stored on-device and act as a highly efficient zero-shot fallback. During inference, evaluating a query embedding against these prototypes incurs negligible computational overhead compared to the final linear classification layer in standard HAR models, as it closely mirrors the matrix multiplication. As demonstrated in our experiments, this imposes essentially no performance penalty compared to the original model, while accommodating future adaptation if calibration data is introduced.

Crucially, this formulation allows existing pretrained deep learning classifiers, such as TinierHAR \cite{bianTinierHARUltraLightweightDeep2025}, to be seamlessly repurposed as ProtoNets. By simply removing the final linear classification layer, the pretrained backbone functions directly as the embedding network within the ProtoNet framework. Although standard ProtoNets are typically optimized using specialized episodic meta-training, prior work \cite{laenen2021episodes} argues that standard training with cross-entropy loss is sufficient to yield a well-clustered embedding space. Our experiments specifically validate this finding for sensor-based HAR. This flexibility allows developers to select from a wide variety of performant, state-of-the-art HAR models and repurpose them without the need for complex retraining from scratch.

Beyond acting as a zero-shot fallback, these prior prototypes preserve
source-domain information that standard ProtoNets discard when adapting
only from support examples. Once computed and stored on-device, they can
therefore additionally serve as regularizers for low-data and noisy calibration.

\subsection{MAP Prototype Estimation}
\label{sec:sup_proto_updating}

Standard ProtoNets estimate each class prototype as the empirical mean of its support embeddings (see \autoref{sec:proto_nets}), which can be interpreted as a Maximum Likelihood Estimate (MLE) fitted only to the support set. To incorporate source-domain information from prior prototypes, we instead formulate prototype estimation as a Bayesian inference problem \cite{murphy2007conjugate} and compute each prototype as Maximum A Posteriori (MAP) estimate:

\begin{equation}
c_k^{\mathrm{MAP}} = \arg\max_{c_k} p(c_k|S_k)
=
\arg\max_{c_k}
\left[ p(S_k|c_k) p(c_k)
\right].
\end{equation}

We model the source-domain prior distribution \(p(c_k)\) and support likelihood \(p(S_k |c_k)\) as Gaussians:

\begin{equation}
    p(c_k)=\mathcal N(c_k|\mu_{D,k},\sigma^2_{D,k})\text{,} \quad p(S_k|c_k)\propto\mathcal N\left(\mu_{S,k}|c_k,\frac{\sigma^2_{S,k}}{N}\right).
    \label{eq:prior}
\end{equation}

Here, $N=|S_k|$, while $\mu_{D,k}$ and $\sigma^2_{D,k}$ denote the empirical mean and variance estimated from the embeddings of the training examples $D_k$, while $\mu_{S,k}$ and $\sigma^2_{S,k}$ are derived from the support examples $S_k$. The variance $\sigma^2_{S,k}/N$ represents the uncertainty of the empirical support mean, so the likelihood becomes more precise as the support size increases. To maintain computational efficiency, we constrain all covariance matrices to be diagonal and represent the corresponding variances as vectors. For $N=1$, the empirical support variance is undefined, so we set $\sigma^2_{S,k}=\sigma^2_{D,k}$.

Because the Gaussian prior and likelihood are conjugate, the posterior $p(c_k |S_k)$ is also Gaussian \cite{murphy2007conjugate}. Its mode coincides with its mean, such that $c_k^{\mathrm{MAP}}=\mu_{\mathrm{post},k}$, yielding the following closed-form expressions:

\begin{equation}
    \frac{1}{\sigma _{\text{post},k}^{2}}=\frac{1}{\sigma^{2}_{D,k}}+\frac{N}{\sigma ^{2}_{S,k}},
    \label{eq:sup_post_var}
\end{equation}

\begin{equation}
    c_k^{\mathrm{MAP}} = \mu _{\text{post},k} = \left( \frac{\sigma_{\text{post},k}^2}{\sigma_{D,k}^2} \right) \mu_{D,k} + \left( \frac{N \sigma_{\text{post},k}^2}{\sigma_{S,k}^2} \right) \mu_{S,k}
     \label{eq:sup_post_mean}
\end{equation}

Intuitively, \autoref{eq:sup_post_mean} can be viewed as a precision-weighted combination of the source-domain prior mean and the target-user support mean. A simpler alternative would be a fixed-weight interpolation:

\begin{equation}
    c_k^{\mathrm{fix}} = (1-\alpha)\mu_{D,k} + \alpha\mu_{S,k} \quad \text{with} \quad \alpha \in [0,1].
\end{equation}

However, this alternative assigns the same trust to the support set regardless of user, activity class, embedding dimension, support size, or calibration noise. It may therefore over-adapt to noisy or unrepresentative support examples, moving prototypes too far away from the source-domain prior prototypes, or under-adapt when support examples are consistent and informative.

In contrast, MAP prototype estimation derives its weights from the relative precisions of the prior and support statistics. This weighting, and thus the degree of adaptation, is class-, \mbox{dimension-,} and data-dependent: when support evidence exhibits statistical noise, the posterior reflects higher epistemic uncertainty about the target prototype and regularizes the estimate toward the prior, improving robustness against outliers, while precise support evidence receives greater influence. Furthermore, this avoids introducing the additional hyperparameter $\alpha$, which would be difficult to tune reliably for unseen users. Since the update is analytical, gradient-free, and efficient to compute, it adds only a low computational overhead on resource-constrained wearable devices (see \autoref{sec:comp_overhead}). The complete step-by-step procedure is outlined in \autoref{alg:bayesian_update}.

\begin{algorithm}[H]
\caption{MAP Prototype Estimation}
\label{alg:bayesian_update}
\begin{algorithmic}[1]
    \STATE Compute support mean $\mu_{S,k}$ and variance $\sigma^2_{S,k}$
    \STATE Compute posterior variance $\sigma_{post,k}^2$ using \autoref{eq:sup_post_var}
    \STATE Compute posterior mean $\mu_{post,k}$ using \autoref{eq:sup_post_mean}
    \RETURN $c_k^\mathrm{MAP}=\mu_{post,k}$
\end{algorithmic}
\end{algorithm}

\subsection{MAP-EM Prototype Estimation}
\label{sec:unsup_proto_updating}

MAP prototype estimation enables robust few-shot adaptation when a labeled support set is available. However, while recording calibration data is feasible in many HAR settings, labeling each sample is often impractical \cite{sheng2025reducing}. We therefore extend MAP prototype estimation to weakly labeled support sets, where only information on which activities were performed is provided. In this setting, we frame the unknown labels $Z=\{z_i\}$ of support embeddings $S=\{s_i\}$ as latent variables. This gives a Gaussian Mixture Model in the embedding space, where each present activity class corresponds to one Gaussian component, and $z_i \in \{1,\dots,K\}$ indicates which component generated $s_i$. We use uniform mixture weights $\pi_k=1/K$ and an efficient fixed isotropic covariance $\Sigma_k=\sigma^2_{\text{EM}}I$. Since these quantities are fixed, the only parameters optimized during adaptation are the component means $\mu_k$, which correspond to updated prototypes $C=\{c_k\}$. To fit the prototypes to the weakly labeled samples, we define the MAP objective: 

\begin{flalign}
    C^\mathrm{MAP}
    &=
    \arg\max_C
    \bigg[
    \underbrace{\log p(S |C)}_{
    \mathclap{\hspace{-0.8cm}
    \displaystyle
    \sum_{i=1}^{N_S}
    \log
    \sum_{k=1}^{K}
    \pi_k
    \mathcal N(s_i |c_k,\sigma^2_{\text{EM}}I)
    }}
    \quad
    + \quad
    \underbrace{\log p(C)}_{
    \mathclap{\hspace{2.6cm}
    \displaystyle
    \sum_{k=1}^{K}
    \log
    \mathcal N(c_k |\mu_{D,k},\sigma^2_{D,k})
    }}
    \bigg]. && \label{eq:map_em_objective}
 \end{flalign}

This objective again incorporates source-domain information using the same Gaussian prototype priors $p(c_k)$ (see \autoref{eq:prior}). Direct optimization is difficult because the class assignments $z_i$ are unobserved, so each support likelihood contains a sum over all possible assignments inside the logarithm. This couples the prototype updates and prevents a simple closed-form solution.

We therefore utilize Expectation-Maximization (EM) \cite{bishop2006pattern}, an iterative optimization algorithm used to estimate model parameters in the presence of latent variables given observed data. The E-step computes the support posterior probabilities, or soft responsibilities, for each support embedding:

\begin{equation}
r_{ik}^{(t)}
=
p(z_i=k |s_i, C^{(t)})
=
\frac{
\pi_k
\mathcal N(s_i |c_k^{(t)}, \sigma^2_{\text{EM}} I)
}{
\sum_{j=1}^{K}
\pi_j
\mathcal N(s_i |c_j^{(t)}, \sigma^2_{\text{EM}} I)
}.
\label{eq:responsibilities}
\end{equation}

Based on these responsibilities, the M-step updates the prototypes using the support evidence together with the prototype prior. First, we compute the soft count and responsibility-weighted support mean for each class:

\begin{equation}
    N_k^{(t)} = \sum_{i=1}^{N_S} r_{ik}^{(t)}
    \quad \quad \text{and} \quad \quad
    \mu_{S,k}^{(t)}
    =
    \frac{1}{N_k^{(t)}}
    \sum_{i=1}^{N_S} r_{ik}^{(t)} s_i .
    \label{eq:soft_count_and_mean}
\end{equation}

The prototype update then is the MAP estimate of the component mean $c_k$ under
the prototype prior $p(c_k)$ and the responsibility-weighted support evidence.
It follows the same closed-form equations as for MAP prototype estimation (see \autoref{eq:sup_post_var}, \autoref{eq:sup_post_mean}), but replaces the labeled support statistics with their soft EM counterparts:

\begin{equation}
    \frac{1}{\sigma_{\text{post},k}^{2(t+1)}}
    =
    \frac{1}{\sigma^2_{D,k}}
    +
    \frac{N_k^{(t)}}{\sigma^2_{\text{EM}}},
    \label{eq:map_em_post_var}
\end{equation}

\begin{equation}
    c_k^{(t+1)} = \mu_{\text{post},k}^{(t+1)} = \left( \frac{\sigma_{\text{post},k}^{2, (t+1)}}{\sigma^2_{D,k}}\right)\mu_{D,k}
        +
        \left(\frac{N_k^{(t)}\sigma_{\text{post},k}^{2, (t+1)}}{\sigma^2_{\text{EM}}}
    \right)\mu_{S,k}^{(t)}.
    \label{eq:map_em_post_mean}
\end{equation}

The soft count $N_k^{(t)}$ acts as the effective number of support examples
assigned to class $k$. Confident responsibilities increase the influence of the
support evidence and move the prototype toward the target-subject support
distribution, while uncertain assignments or small soft counts leave the prior
prototype mean $\mu_{D,k}$ with greater influence. After a fixed number of EM
iterations, the final prototype estimates are $C^{(T)}=\{c_k^{(T)}\}$.

Before applying MAP-EM prototype estimation and final query classification, we center the embedding space to reduce subject-specific distribution shifts. For the train set, this can be viewed as centering each prototype with the global train embedding mean $\bar d$, and support and query embeddings with the support mean $\bar s$:

\begin{equation}
    \tilde{\mu}_{D,k}=\mu_{D,k}-\bar d, \quad\quad \tilde s_i=s_i-\bar s, \quad \quad \tilde q_i=q_i-\bar s.
    \label{eq:centering}
\end{equation}

This removes global subject-specific offsets that affect all activities in a similar direction, allowing MAP-EM to focus on class-specific changes in the target subject rather than compensating for a global shift that can be easily corrected by centering. Overall, the complete step-by-step procedure is outlined in \autoref{alg:map_em_update}.

\begin{algorithm}[H]
\caption{MAP-EM Prototype Estimation}
\label{alg:map_em_update}
\begin{algorithmic}[1]
    \STATE Center support embeddings $S$ using \autoref{eq:centering}
    \STATE Initialize prototypes $c_k^{(0)}=\tilde{\mu}_{D,k}$
    \FOR{$t = 0 \dots T-1$}
        \STATE Compute responsibilities $r_{ik}^{(t)}$ using \autoref{eq:responsibilities}
        \STATE Compute soft count $N_k^{(t)}$, support mean $\mu_{S,k}^{(t)}$ using \autoref{eq:soft_count_and_mean}
        \STATE Compute posterior variance $\sigma_{\text{post},k}^{2(t+1)}$ using \autoref{eq:map_em_post_var}
        \STATE Compute posterior mean $\mu_{\text{post},k}^{(t+1)}=c_k^{(t+1)}$ using \autoref{eq:map_em_post_mean}
    \ENDFOR
    \RETURN $C^{(T)}=\{c_k^{(T)}\}$
\end{algorithmic}
\end{algorithm}

\vspace*{-.5em}

\section{Evaluation}

\begin{figure*}[t]
    \centering
    \includegraphics[width=\linewidth]{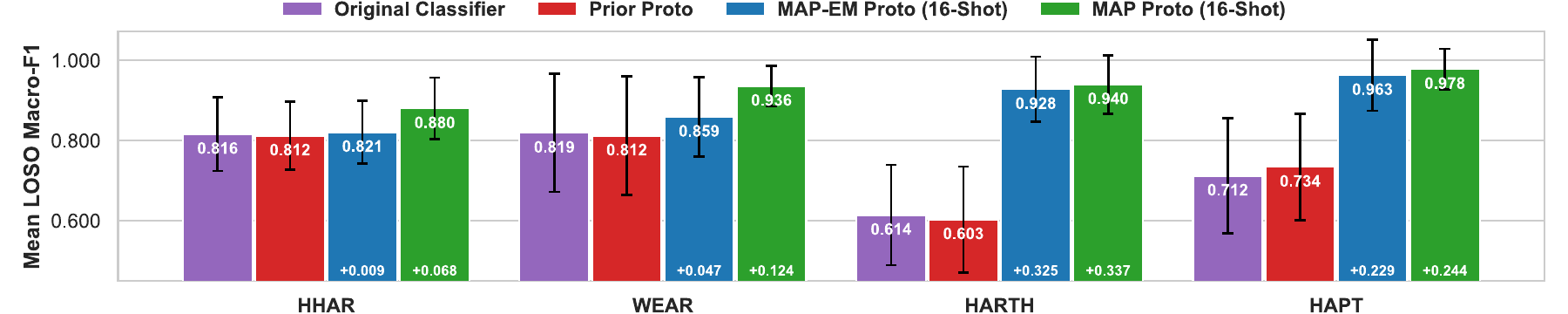} 
    \caption{Mean LOSO macro-F1 across four HAR datasets for the original classifier, its repurposing with prior prototypes, and our proposed weakly supervised (MAP-EM Proto) and supervised (MAP Proto) adaptation methods  on 16 shots per class.}
    \Description{TODO}
    \label{fig:bars}
\end{figure*}

\begin{table}[b]
\centering
\caption{HAR datasets used for benchmarking.}
\label{table:datasets}
\small
\begin{tabular}{lllll}
\toprule
Dataset & Subjects & Activities & Channels & Sensors \\
\midrule
HHAR \cite{stisen_smart_2015} & 9 & 6 & 6 & Acc, Gyr  \\
WEAR \cite{bockWEAROutdoorSports2024} & 24 & 18 & 12 & Acc  \\
HARTH \cite{logacjov_harth_2021} & 22 & 12 & 6 & Acc  \\
HAPT \cite{reyes-ortiz_transition-aware_2016} & 30 & 12 & 6 & Acc, Gyr \\
\bottomrule
\end{tabular}
\end{table}

We evaluate on four HAR benchmark datasets (see \autoref{table:datasets}) using the WHAR Datasets library \cite{burzer2025whar} with 3-second windows and leave-one-subject-out (LOSO) cross-validation. For each fold, the held-out subject is used exclusively for testing, while the remaining subjects are split into 80\% training and 20\% validation data. We use the efficient TinierHAR \cite{bianTinierHARUltraLightweightDeep2025} as base classifier and train each model using AdamW \cite{loshchilov2017decoupled} with cross-entropy loss for up to 100 epochs, batch size 64, learning rate $10^{-4}$, 50\% window overlap, and early stopping on validation macro-F1 with patience 10 for model selection. We report mean LOSO macro-F1 and standard deviation. For few-shot evaluation, we disable window overlap and average results over 100 episodes for 1 to 16 shots per class. Support and query sets are strictly disjoint, and all samples not selected for support are used as queries. For MAP-EM, we use one iteration with $\sigma_{EM}^{2}=0.5$.

\begin{figure}[h]
    \centering
    \includegraphics[width=0.9\linewidth]{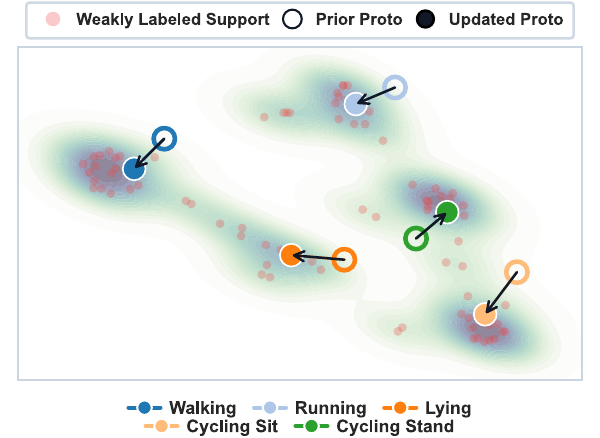} 
    \caption{T-SNE \cite{van2008visualizing} visualization of one MAP-EM update on HARTH with 16 weakly labeled support examples per class.}
    \Description{TODO}
    \label{fig:tsne}
\end{figure}

As shown in \autoref{fig:bars}, prior prototypes successfully repurpose the pretrained classifier as a ProtoNet while preserving zero-shot performance. With 16 labeled shots (48 seconds of calibration) per activity, MAP prototype estimation yields substantial test macro-F1 gains across all datasets, ranging from \(+6.8\) pp on HHAR to \(+33.7\) pp on HARTH. MAP-EM achieves comparable gains, ranging from \(+0.9\) pp on HHAR to \(+32.5\) pp on HARTH. \autoref{fig:tsne} illustrates this behavior via T-SNE~\cite{van2008visualizing}: the support embeddings are initially shifted away from the prior prototypes due to domain shift, while a single MAP-EM iteration moves the prototypes toward the target-subject clusters and repairs the decision boundaries. Both methods also visibly reduce LOSO variance compared to the zero-shot baseline.

\begin{figure*}[t]
    \centering
    \includegraphics[width=\linewidth]{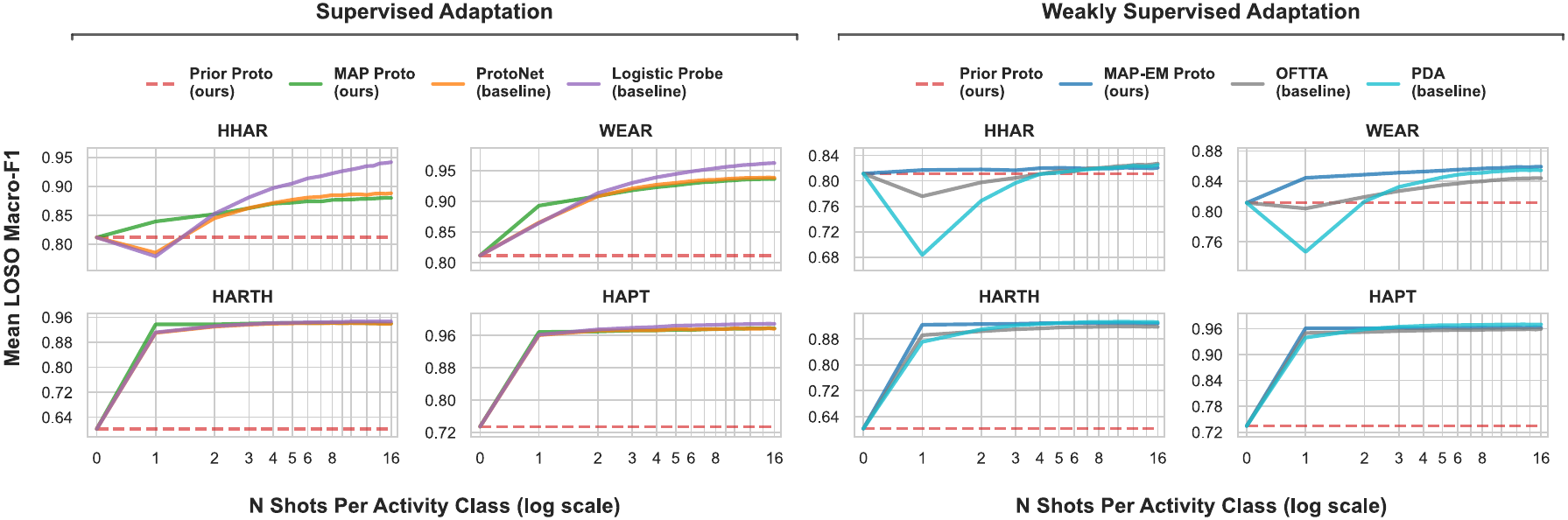} 
    \caption{Mean LOSO Macro-F1 across support set sizes for supervised and weakly supervised adaptation baselines.}
    \Description{TODO}
    \label{fig:curve_sup}
\end{figure*}

\autoref{fig:curve_sup} illustrates how macro-F1 performance scales with number of support examples per class. We compare against several baselines, all based on the same pretrained classifier backbones. For the supervised setting, we compare against standard ProtoNets \cite{snell2017prototypical} and a logistic regression probe fitted to the calibration data. The latter is gradient-based and therefore unsuitable for on-device adaptation, but serves as an approximate performance ceiling. 
For the weakly supervised setting, we compare against weakly supervised variants of PDA \cite{bohdalFeedForwardSourceFreeDomain2023}, which builds prototypes from pseudo-labeled support data, and OFTTA \cite{wangOptimizationFreeTestTimeAdaptation2024}, a HAR-specific TTA method that we adapted to the offline setting by freezing updates after calibration.

Looking closely at the curves in \autoref{fig:curve_sup}, adaptation performance largely converges around 16 support shots per class across all methods. At these higher support sizes, our proposed approaches match the peak performance of the established baselines. The logistic probe establishes the overall performance ceiling on HHAR, WEAR, and HAPT by leveraging gradients. However, stark differences emerge in the extreme low-shot regime of just one shot per class (only 3 seconds). \autoref{tab:k1_gains} reports the corresponding macro-F1 gains over the prior-prototype zero-shot baseline. Both proposed methods improve from the first shot on all datasets: MAP gains between \(+2.76\) pp on HHAR and \(+33.44\) pp on HARTH, while MAP-EM remains consistently positive with gains from \(+0.56\) pp to \(+32.13\) pp. In contrast, the supervised baselines are less reliable, with ProtoNets and the logistic probe falling below the zero-shot baseline on HHAR by \(-2.65\) pp and \(-3.27\) pp, respectively. This vulnerability is even more pronounced in the weakly supervised setting: OFTTA and PDA collapse below the baseline on both HHAR and WEAR, with PDA dropping by \(-12.81\) pp on HHAR and \(-6.48\) pp on WEAR. Even where these baselines still improve, MAP-EM achieves the strongest one-shot performance, highlighting its ability to avoid early-stage degradation in the ultra-low-shot regime.

\begin{table}[h]
\centering
\caption{Macro-F1 gains over the prior-prototype zero-shot baseline at one shot (3 seconds of calibration) per activity.}
\label{tab:k1_gains}
\small
\setlength{\tabcolsep}{3.5pt}
\begin{tabular}{l rrr rrr}
\toprule
& \multicolumn{3}{c}{Supervised} & \multicolumn{3}{c}{Weakly Supervised} \\
\cmidrule(lr){2-4}\cmidrule(lr){5-7}
Dataset & MAP & ProtoNet & Log. Probe & MAP-EM & OFTTA & PDA \\
\midrule
HAPT  & \textbf{+23.42} & +22.63 & +22.74 & \textbf{+22.67} & +21.55 & +20.49 \\
HARTH & \textbf{+33.44} & +30.73 & +30.88 & \textbf{+32.13} & +28.91 & +26.84 \\
HHAR  & \textbf{+2.76}  & -2.65  & -3.27  & \textbf{+0.56}  & -3.57  & -12.81 \\
WEAR  & \textbf{+8.12}  & +5.37  & +5.26  & \textbf{+3.29}  & -0.76  & -6.48 \\
\bottomrule
\end{tabular}
\end{table}

\vspace*{-.5em}

\subsection{Computational Overhead}
\label{sec:comp_overhead}

We quantify only the prototype estimation cost after embedding
extraction. MAP scales roughly as $\mathcal{O}((N+K)d)$, and MAP-EM as
$\mathcal{O}(TNKd)$, with $K$ adapted classes, $N$ support
embeddings of dimension $d$, and $T$ EM steps. We measure the
concrete values shown in \autoref{tab:adaptation_flops} with an
operation-level counter matching our implementation. Across 1 to 16 shots, MAP requires
only $0.3$--$3.9\%$ of a single TinierHAR backbone forward pass. MAP-EM
is more expensive, but also remains below one pass, requiring
$1.4$--$82.4\%$. This low cost is enabled by diagonal covariance in MAP
and isotropic covariance in MAP-EM: although full covariances could model
correlations between embedding dimensions, they would require estimating,
storing, and inverting $d \times d$ matrices, whereas our restrictions
reduce adaptation to lightweight vector operations.

\begin{table}[h]
\small
\centering
\caption{Measured cost (excluding embedding extraction) in MFLOPs with $d=32$, $T=1$, and all classes adapted.}
\label{tab:adaptation_flops}
\begin{tabular}{lrrrrrr}
\toprule
Dataset & $K$ & \multicolumn{2}{c}{1-shot} & \multicolumn{2}{c}{16-shot} & TinierHAR \\
\cmidrule(lr){3-4}\cmidrule(lr){5-6}
& & MAP & MAP-EM & MAP & MAP-EM & \\
\midrule
HHAR  & 6  & 0.002 & 0.008 & 0.014 & 0.104 & \textbf{0.586} \\
HAPT  & 12 & 0.003 & 0.028 & 0.028 & 0.400 & \textbf{0.750} \\
HARTH & 12 & 0.003 & 0.028 & 0.028 & 0.400 & \textbf{0.870} \\
WEAR  & 18 & 0.005 & 0.060 & 0.042 & 0.889 & \textbf{1.079} \\
\bottomrule
\end{tabular}
\end{table}

\vspace*{-.5em}

\section{Discussion} 

The one-shot results demonstrate the instability of purely empirical adaptation under extremely limited calibration. Standard ProtoNets, logistic probing, OFTTA, and PDA each fall below the prior-prototype zero-shot baseline on at least one dataset, indicating sensitivity to noisy support estimates or unreliable pseudo-labels. In contrast, MAP and MAP-EM consistently improve upon the baseline using only one labeled or weakly labeled support example (3 seconds of data) per class. By combining target-user evidence with source-domain prototype priors, they reduce the risk of low-data degradation. As the support set grows to 16 shots per class, target-domain statistics become more reliable, the influence of the prior decreases, and empirical baselines approach the proposed methods. This is the intended behavior of the prior: it regularizes uncertain estimates in the ultra-low-shot regime while progressively yielding to user-specific evidence. Although logistic probing performs strongly with sufficient labeled data, its reliance on labels and gradient-based optimization limits its suitability for lightweight on-device adaptation. In contrast, our framework provides gradient-free adaptation across zero-shot, supervised, and weakly supervised calibration settings, with low computational overhead after embedding extraction. This combination of robustness, flexibility, and efficiency makes it well suited to resource-constrained wearable devices.

Several limitations remain. First, MAP-EM is currently weakly rather than fully unsupervised, since it requires knowledge of which activity classes were performed during calibration to determine which GMM components to adapt. Future work should infer the activity classes directly from the calibration data, enabling fully unsupervised adaptation. Second, although adaptation itself is source-free, repurposing a pretrained HAR classifier is not, because the prior prototypes are obtained through a one-time offline pass over the original training data. Deriving equivalent priors directly from the pretrained classifier weights would remove this dependency. Third, MAP and MAP-EM use uncertainty estimates to regularize adaptation, but they remain plug-in prototype estimators that use the ProtoNet decision rule for prediction. Future work should extend the framework to uncertainty-aware prediction through full Bayesian inference over the predictive distribution.

\section{Conclusion}

We presented BayaHAR, a lightweight, gradient-free framework for user adaptation in HAR on resource-constrained wearable devices. By repurposing pretrained classifiers as ProtoNets and updating their prototypes with MAP and MAP-EM, BayaHAR supports zero-shot, supervised few-shot, and weakly supervised few-shot adaptation within a single model. Across four HAR datasets, prior prototypes preserve zero-shot performance, while both methods already improve reliably from one shot (3 seconds of calibration) per activity, avoiding the early-stage degradation observed in empirical baselines. With 16 shots (48 seconds) per activity, MAP achieves macro-F1 gains of up to \(+33.7\) pp, while MAP-EM reaches up to \(+32.5\) pp. Since adaptation requires only closed-form prototype updates, with MAP-EM using a single EM step, it is well suited for practical on-device personalization of preexisting HAR classifiers.

\begin{acks}

This work was supported by the IPAI Foundation gGmbH (Science Residency Program) and the HammerHAI project, which received funding from the EuroHPC JU (grant No. 101234027) co-funded by the European Commission, the BMFTR, and the State Ministries of Science of Baden-Württemberg, Bavaria, and Lower Saxony.

\end{acks}

\bibliographystyle{ACM-Reference-Format}
\balance
\bibliography{bibliography}

\end{document}